\documentclass[11pt]{article}
\usepackage{coling2020}
\usepackage{times}
\usepackage{url}
\usepackage{latexsym}

\usepackage{tikz}
\usepackage{subcaption}
\usepackage{multirow}
\usepackage{algorithm}
\usepackage{algorithmicx}
\usepackage{algpseudocode}
\usepackage{booktabs}
\usepackage{amssymb}
\usepackage{pifont}
\usepackage{todonotes}
\usepackage{amsfonts}
\usepackage{amsmath}
\newcommand{\cmark}{\ding{51}}
\newcommand{\xmark}{\ding{55}}

\colingfinalcopy 

\title{CoLAKE: Contextualized Language and Knowledge Embedding}

\author{
  Tianxiang Sun\textsuperscript{1,}\thanks{{} {} Work done during internship at Amazon Shanghai AI Lab.} , Yunfan Shao\textsuperscript{1}, Xipeng Qiu\textsuperscript{1,}\thanks{{} {} Corresponding author.}\\ 
  \bf Qipeng Guo\textsuperscript{1}, Yaru Hu\textsuperscript{1}, Xuanjing Huang\textsuperscript{1}, Zheng Zhang\textsuperscript{2}\\
  \textsuperscript{1}Shanghai Key Laboratory of Intelligent Information Processing, Fudan University \\
  \textsuperscript{1}School of Computer Science, Fudan University \\
  \textsuperscript{2}Amazon Shanghai AI Lab \\
  {\tt \{txsun19,yfshao19,xpqiu,qpguo16,xjhuang\}@fudan.edu.cn}\\
  {\tt yrhu112358@outlook.com zhaz@amazon.com}\\}

\date{}

\begin{document}
\maketitle
\begin{abstract}
  With the emerging branch of incorporating factual knowledge into pre-trained language models such as BERT, most existing models consider shallow, static, and separately pre-trained entity embeddings, which limits the performance gains of these models. Few works explore the potential of deep contextualized knowledge representation when injecting knowledge. In this paper, we propose the Contextualized Language and Knowledge Embedding (CoLAKE), which jointly learns contextualized representation for both language and knowledge with the extended MLM objective. Instead of injecting only entity embeddings, CoLAKE extracts the knowledge context of an entity from large-scale knowledge bases. To handle the heterogeneity of knowledge context and language context, we integrate them in a unified data structure, word-knowledge graph (WK graph). CoLAKE is pre-trained on large-scale WK graphs with the modified Transformer encoder. We conduct experiments on knowledge-driven tasks, knowledge probing tasks, and language understanding tasks. Experimental results show that CoLAKE outperforms previous counterparts on most of the tasks. Besides, CoLAKE achieves surprisingly high performance on our synthetic task called word-knowledge graph completion, which shows the superiority of simultaneously contextualizing language and knowledge representation.\footnote{Our code is available at \url{https://github.com/txsun1997/CoLAKE}.}
\end{abstract}

\section{Introduction}

Deep contextualized language models pre-trained on large-scale unlabeled corpora have achieved significant improvement on a wide range of NLP tasks~\cite{Peters2018Deep,Devlin2019BERT,Yang2019XLNet}. However, they are shown to have difficulty capturing factual knowledge~\cite{Logan2019Barack}.

Recently, there is a growing interest in combining pre-trained language models (PLMs) with structured, human-curated knowledge. A popular approach is to inject pre-trained entity embeddings into PLMs to better capture factual knowledge, such as ERNIE~\cite{Zhang2019ERNIE} and KnowBERT~\cite{Peters2019Know}. The shortcomings of these models can be summarized as follows: (1) The entity embeddings are separately pre-trained with some knowledge embedding (KE) models (e.g., TransE~\cite{Bordes2013TransE}), and fixed during training PLMs. Thus they are not real joint models to learn the knowledge embedding and language embedding simultaneously.   (2) The previous models only take entity embeddings to enhance PLMs, which are hard to fully capture the rich contextual information of an entity in the knowledge graph (KG). Thus their performance gains are limited by the quality of pre-trained entity embeddings. (3) The pre-trained entity embeddings are static and need to be re-trained when the KG is slightly changed, e.g., a new entity is added.



\begin{figure}[t]
  \centering
  \includegraphics[width=.95\columnwidth]{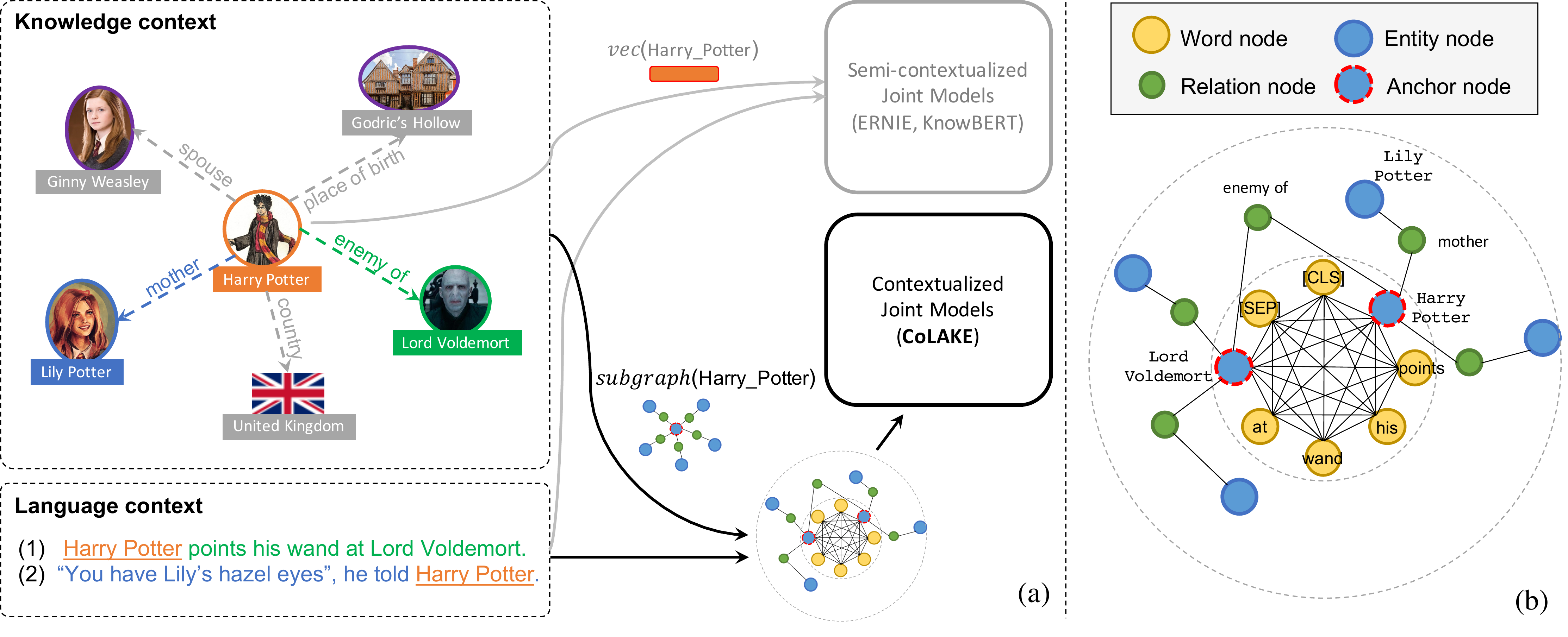}
  \caption{(a) When injecting knowledge, CoLAKE considers the \textit{knowledge context} about the entity while previous semi-contextualized joint models only consider a single entity embedding. By contextualizing the entity, CoLAKE is able to directly access (\texttt{Harry\_Potter}, \emph{enemy of}, \texttt{Lord\_Voldemort}) to help understand sentence (1) and access (\texttt{Harry\_Potter}, \emph{mother}, \texttt{Lily\_Poter}) to help understand sentence (2). (b) The word-knowledge graph (WK graph) is a unified structure to represent both the language context and the knowledge context, which is composed of two parts: the fully-connected word graph (indicated by the inner dashed circle) and the knowledge sub-graphs extracted from the large KGs (the outer dashed circle).}
  \label{fig:example}
\end{figure}

The above observation motivates us to explore a better model to integrate language representation and factual knowledge. In this paper, we propose the \textbf{Co}ntextualized \textbf{L}anguage \textbf{A}nd \textbf{K}nowledge \textbf{E}mbedding (\textbf{CoLAKE}), which jointly learns language representation and knowledge representation in a common representation space. Different from the previous models, CoLAKE dynamically represents an entity according to its \textit{knowledge context} and \textit{language context}. For each entity, CoLAKE considers a sub-graph surrounding it as its knowledge context that contains the facts (triplets) about the entity. In this way, CoLAKE can dynamically access different facts as background knowledge to help understand the current text. As shown in Figure~\ref{fig:example}(a), to understand different sentences, CoLAKE can utilize different facts about the linked entity \texttt{Harry\_Potter}. The \emph{knowledge context} of \texttt{Harry\_Potter} is a sub-graph containing the triplets about it. According to whether or not to utilize entities' knowledge context, our proposed CoLAKE can be distinguished from previous models, which we call \textit{semi-contextualized joint models } since they only contextualize language representation.


To deal with the heterogeneous structure of language and KG, we build a graph to integrate them into a unified data structure, called \emph{word-knowledge graph (WK graph)}. Most recent successful PLMs use Transformer architecture~\cite{Vaswani2017Attention}, which treats input sequences as fully-connected word graphs. WK graph is knowledge-augmented word graph. Using entities mentioned in the sentence, we extract sub-graphs centered on those mentioned entities from KGs. Then we mosaic such sub-graphs and the word graph in a unified heterogeneous graph, i.e. WK graph. An instance of the WK graph can be found in Figure~\ref{fig:example}(b). The constructed WK graph is fed into CoLAKE along with its adjacency matrix to control the information flow to reflect the graph structure. CoLAKE is based on the Transformer encoder, with the embedding layer and the encoder layers slightly modified to adapt to input in the form of WK graph. Besides, we extend the masked language model (MLM) objective~\cite{Devlin2019BERT} to the whole input graph. That is, apply the same masking strategy to word, entity, and relation nodes and training the model to predict the masked nodes based on the rest of the graph.


We evaluate CoLAKE on several knowledge-required tasks and GLUE~\cite{Wang2019GLUE}. Experimental results demonstrate that CoLAKE outperforms previous semi-contextualized counterparts on most of the tasks. To explore potential applications of CoLAKE, we design a synthetic task called word-knowledge graph completion. Our evaluation on this task shows that CoLAKE outperforms several KE models by a large margin, in transductive setting and inductive setting.

In summary, CoLAKE can be characterized in three-fold: (1) CoLAKE learns contextualized language representation and contextualized knowledge representation simultaneously with the extended MLM objective. (2) CoLAKE adopts the WK graph to integrate the heterogeneous input for language and knowledge. (3) CoLAKE is essentially a pre-trained graph neural network (GNN),
thereby being structure-aware and easy to extend.


\section{Related Work}

\paragraph{Language Representation Learning.} The past decade has witnessed the great success of pre-trained language representation. Initially, word representation pre-trained using multi-task objectives~\cite{Collobert2008Unified} or co-occurrence statistics~\cite{Mikolov2013word2vec,Pennington2014Glove} are static and non-contextual. Recently, contextualized word representation pre-trained on large-scale unlabeled corpora with deep neural networks has dominated across a wide range of NLP tasks~\cite{Peters2018Deep,Devlin2019BERT,Yang2019XLNet,Qiu2020PTM}.

\paragraph{Knowledge Representation Learning.} Knowledge Representation Learning (KRL) is also termed as Knowledge Embedding (KE), which is to map entities and relations into low-dimensional continuous vectors. Most existing methods use triplets as training samples to learn static, non-contextual embeddings for entities and relations~\cite{Bordes2013TransE,Yang2015DistMult,Lin2015TransR}. Recent advances focusing on contextualized representation, which use subgraphs or paths as training samples, have achieved new state-of-the-art results on KG tasks~\cite{Wang2019CoKE,Wang2020Entity}.


\paragraph{Joint Language and Knowledge Models.} Due to the mutual information existing in language and KGs, joint models often benefit both sides. Besides, tasks such as entity linking also require entity embeddings that are compatible with word embeddings. Combining the success of Mikolov et al.~\shortcite{Mikolov2013word2vec} and Bordes et al.~\shortcite{Bordes2013TransE}, Wang et al.~\shortcite{Wang2014Knowledge} jointly learn embeddings for language and KG. Targeting mention-entity matching in entity linking, Yamada et al.~\shortcite{Yamada2016Joint}, Ganea and Hofmann~\shortcite{Ganea2017Joint} also proposed joint methods to map entities and words into the same vector space. Inspired by the recent success of contextualized language representation, much effort has been devoted to injecting entity embeddings into PLMs~\cite{Zhang2019ERNIE,Peters2019Know}. Despite their success, the knowledge gains are limited by the expressivity of their used pre-trained entity embeddings, which is static and inflexible. In contrast, KEPLER~\cite{Wang2019KEPLER} aims to benefit both sides so jointly learn language model and knowledge embedding. However, KEPLER does not directly learn embeddings for each entity but learns to generate entity embeddings with PLMs from entity descriptions. Besides, none of these work exploits the potential of contextualized knowledge representation, which makes them different from our proposed CoLAKE. A brief comparison can be found in Table~\ref{tab:comp}. CoLAKE is conceptually similar to K-BERT~\cite{Liu2020KBERT} and BERT-MK~\cite{He2019Integrating}. CoLAKE differs from K-BERT in that, instead of injecting triplets during fine-tuning, CoLAKE jointly learns embeddings for entities and relations during pre-training LMs. Besides, CoLAKE places language and knowledge representation learning into a unified pre-training task, masked language model, which makes it more concise than BERT-MK. In addition, CoLAKE is a general-purpose joint model while BERT-MK mainly focuses on medical domain.

\begin{table}[htbp]
  \centering\small
  \begin{tabular}{llcccc}
    \toprule
    \multirow{2}{*}{}                    & \multirow{2}{*}{Joint Models}                                   & \multicolumn{2}{c}{Language}      & \multicolumn{2}{c}{Knowledge}      \\
                                         &                                                                 & Objective & Contextualized?       & Objective  & Contextualized?       \\ 
                                         \midrule
    \multirow{2}{*}{Non-contextual}      & Wang et al.~\shortcite{Wang2014Knowledge} & Skip-Gram & \xmark & TransE     & \xmark \\
                                         & Yamada et al.~\shortcite{Yamada2016Joint} & Skip-Gram & \xmark & Skip-Gram  & \xmark \\ \midrule
    \multirow{3}{*}{Semi-contextualized} & ERNIE~\cite{Zhang2019ERNIE}               & MLM       & \cmark & TransE$^*$ & \xmark \\
                                         & KnowBERT~\cite{Peters2019Know}            & MLM       & \cmark & -          & \xmark \\
                                         & KEPLER~\cite{Wang2019KEPLER}              & MLM       & \cmark & TransE     & \xmark \\ \midrule
    Contextualized                       & CoLAKE (Ours)                             & MLM       & \cmark & MLM        & \cmark \\ \bottomrule
    \end{tabular}
  \caption{Comparison of several joint models based on whether the representation is contextualized. $^*$The entity embeddings are fixed during pre-training ERNIE. KnowBERT does not have restrictions on the entity embeddings. For ERNIE and KnowBERT, we omit the next sentence prediction (NSP) objective.}
  \label{tab:comp}
\end{table}

\section{CoLAKE}
\label{sec:model}
\begin{figure}[htb]
  \centering
  \includegraphics[width=\columnwidth]{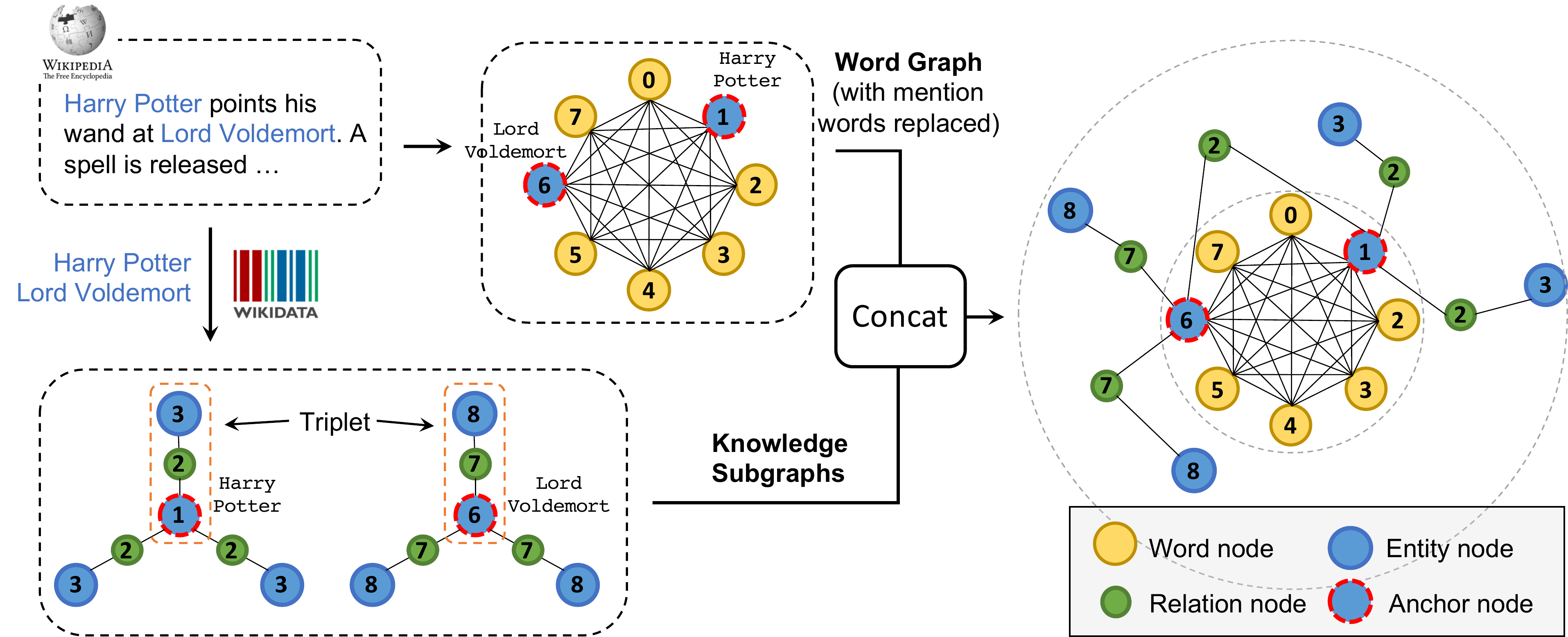}
  \caption{Illustration of WK graph construction. The WK graph is an undirected heterogeneous graph. The numbers marked on graph nodes indicate the position index introduced in Section~\ref{sec:embed}.}
  \label{fig:graph_cons}
\end{figure}

CoLAKE jointly learns contextualized representation for language and knowledge by pre-training on structured, unlabeled \emph{word-knowledge graphs (WK graphs)}. We first introduce how to construct such WK graphs, then we describe the model architecture and the implementation details.

\subsection{Graph Construction}
Typically, language embedding models take sequences as input while knowledge embedding (KE) models take triplets or knowledge sub-graphs as input. Recent successful PLMs take Transformer~\cite{Vaswani2017Attention} as their backbone architecture, which actually processes sequences as fully-connected word graphs. Thus, graph is a common data structure to represent language and knowledge. In this section, we show how to integrate word graph and knowledge sub-graphs into the unified WK graph.

We first tokenize a sentence into a sequence of tokens and fully connect them as a word graph. Then we recognize the mentions in the sentence and use an entity linker to find the corresponding entities in a certain KG. The mention nodes are then replaced by their linked entities, which are called \emph{anchor nodes}. By this replacement, the model is encouraged to map the injected entities and mention words near one another in the vector space. Centered on the anchor nodes $\{e_i\}_i$, we can extract their knowledge contexts $\{\{e_i, r_{ij}, e_{ij}\}_j\}_i$ to form sub-graphs, in which relations are also transformed into graph nodes. The extracted sub-graphs and the word graph are then concatenated with anchor nodes to obtain the WK graph. Figure~\ref{fig:graph_cons} shows the process of constructing a WK graph. In practice, for each anchor node we randomly select up to 15 neighboring relations and entities to construct a sub-graph to be injected into the WK graph. We only consider triplets in which anchor node is head (subject) instead of tail (object). In the WK graph, entities are unique but relations are allowed to repeat.

\subsection{Model Architecture}
\begin{figure}[h]
  \centering
  \includegraphics[width=0.9\columnwidth]{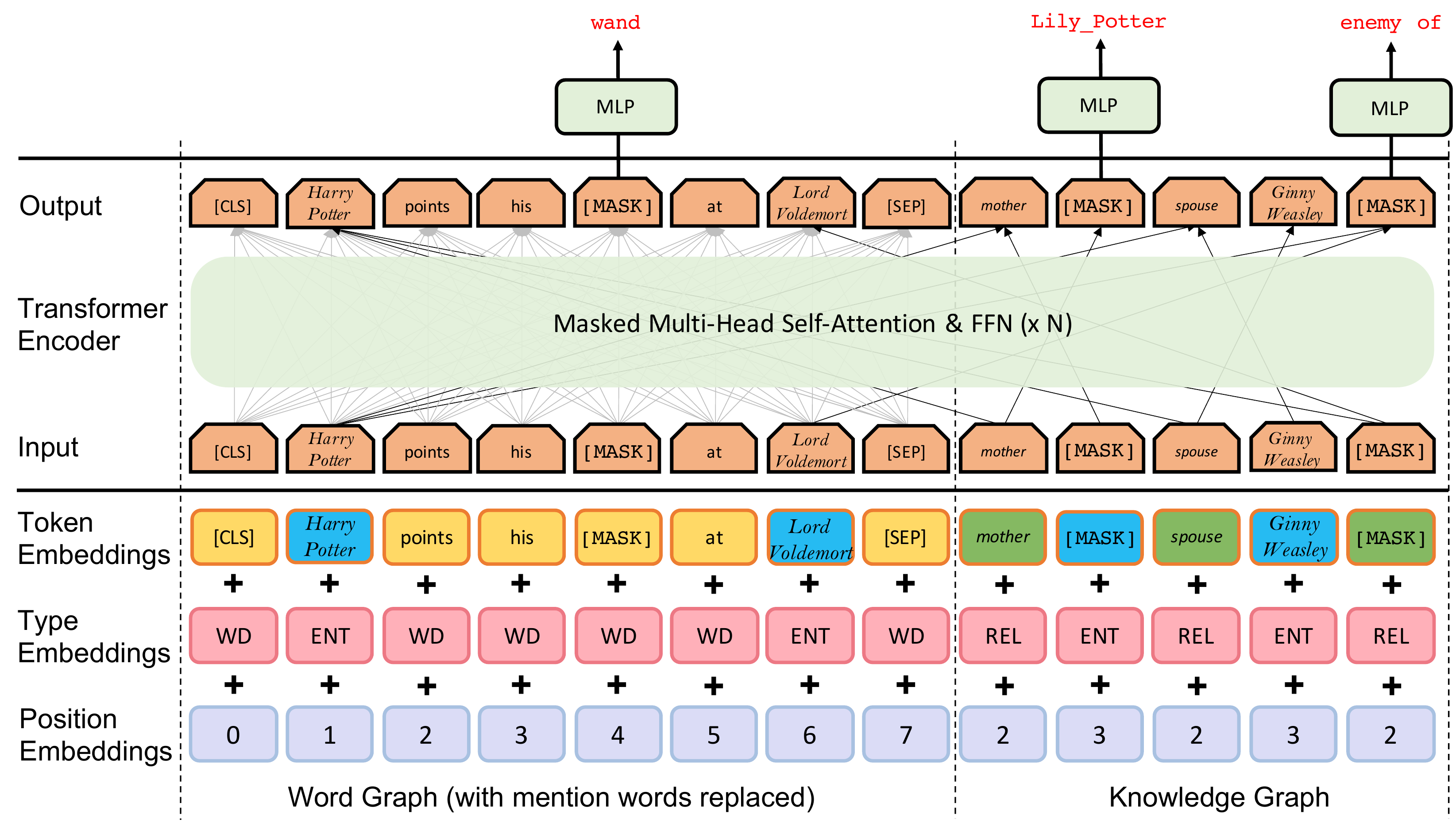}
  \caption{Overall architecture of CoLAKE. In this case, three triplets, (\texttt{Harry\_Potter}, \emph{mother}, \texttt{Lily\_Potter}), (\texttt{Harry\_Potter}, \emph{spouse}, \texttt{Ginny\_Weasley}), and (\texttt{Harry\_Potter}, \emph{enemy of}, \texttt{Lord\_Voldemort}) are injected into the raw sequence. The model is asked to predict the masked word \emph{wand}, the masked entity \texttt{Lily\_Potter}, and the masked relation \emph{enemy of}.}
  \label{fig:archi}
\end{figure}

The constructed WK graphs are then fed into the Transformer~\cite{Vaswani2017Attention} encoder. We modify the embedding and encoder layers of vanilla Transformer to adapt to input in the form of WK graph.

\paragraph{Embedding Layer.}
\label{sec:embed}
The input embedding is the sum of token embedding, type embedding, and position embedding. For token embedding, we maintain three lookup tables for words, entities, and relations respectively. For word embedding, we follow RoBERTa~\cite{Liu2019Roberta} which uses Byte-Pair Encoding (BPE)~\cite{Sennrich2016NMT} to transform sequence into subwords units to handle the large vocabulary. In contrast, we directly learn embeddings for each unique entity and relation as common knowledge embedding methods do. The token embeddings are obtained by concatenating word, entity, and relation embeddings, which are of the same dimensionality. There are different types of nodes so the WK graph is heterogeneous. To handle this, we simply use type embedding to indicate the node types, i.e. word, entity, and relation. For position embedding, we need to assign each injected entity and relation a position index. Inspired by Liu et al.~\shortcite{Liu2020KBERT}, we adopt the soft-position index which allows repeated position indices and keeps tokens in the same triplet continuous. Figure~\ref{fig:graph_cons} shows an intuitive example of how to assign position index to graph nodes.


\paragraph{Masked Transformer Encoder.}
We use masked multi-head self-attention to control the information flow to reflect the structure of WK graph. Given the representation of graph nodes $\mathbf{X}\in \mathbb{R}^{n\times d}$, where $n$ is the number of nodes and $d$ is the dimension for each node, the representation after masked self-attention is obtained by
\begin{align}
	\mathbf{Q}, \mathbf{K}, \mathbf{V} &= \mathbf{X}\mathbf{W}^Q, \mathbf{X}\mathbf{W}^K, \mathbf{X}\mathbf{W}^V,\\
	\mathbf{A} &= \frac{\mathbf{Q}\mathbf{K}^\top}{\sqrt{d_k}}, \label{eq:attn}\\
	\mathrm{Attn}(\mathbf{Q},\mathbf{K},\mathbf{V}) &= \mathrm{Softmax}(\mathbf{A} + \mathbf{M})\mathbf{V},
\end{align}
where $\mathbf{W}^Q, \mathbf{W}^K, \mathbf{W}^V\in \mathbb{R}^{d\times d_k}$ are learnable parameters. $\mathbf{M}\in \mathbb{R}^{n\times n}$ is the mask matrix given by
\begin{equation}
  \mathbf{M}_{ij} =
  \begin{cases}
    0 & \mbox{if $x_i$ and $x_j$ are connected,}\\
    -\inf & \mbox{if $x_i$ and $x_j$ are disconnected.}
  \end{cases}
\end{equation}

With the masked Transformer encoder, each node can only gather information from its 1-hop neighbor at each layer. Masked Transformer encoder works similar to GAT~\cite{Velickovic2018GAT}.

\subsection{Pre-Training Objective}
\label{subsec:mlm}
The Masked Language Model (MLM) objective is to randomly mask some of tokens from the input and train the model to predict the original vocabulary id of the masked tokens based on their contexts. In this section, we extend the MLM from word sequences to WK graphs.

In particular, we mask 15\% of graph nodes at random. When a node is masked, we replace it with (1) the \texttt{[MASK]} token 80\% of time, (2) a randomly sampled node with the same type as the original node 10\% of time, (3) the unchanged node 10\% of time. As different types of nodes are masked, we encourage CoLAKE to learn different aspects of capabilities:
\begin{itemize}
  \item \textbf{Masking word nodes.} When words are masked, the objective is similar to traditional MLM. The difference is, CoLAKE can predict masked words based on not only the context words but also the entities and relations in the WK graph. Masking words helps CoLAKE learn linguistic knowledge.
  \item \textbf{Masking entity nodes.} If the masked entity is an anchor node, the objective, which is to predict the anchor node based on its context, helps to align the representation spaces of language and knowledge. Take the instance in Figure~\ref{fig:example}(b), the embedding of entity \texttt{Harry\_Potter} will be learned to be similar to its textual form, \emph{Harry Potter}. If the masked entity is not an anchor node, the MLM objective is similar to that used in semantic matching-based KE methods such as ConvE~\cite{Dettmers2018ConvE} and CoKE~\cite{Wang2019CoKE}, which enables CoLAKE to learn a large number of entity embeddings. Masking entity nodes helps CoLAKE (a) map words and entities into a common representation space, and (b) learn contextualized representation for entities.
  \item \textbf{Masking relation nodes.} If the masked relation is between two unique anchor nodes, the objective is similar to distantly supervised relation extraction~\cite{Craven1999Constructing}, which requires the model to classify the relationship between two entities mentioned in the text. Otherwise, the objective is to predict the relationship between its two neighboring entities, which is similar to traditional KE methods. Masking relation nodes helps CoLAKE (a) learn to do relation extraction, and (b) learn contextualized representation for relations.
\end{itemize}

However, the pre-training task of predicting masked anchor nodes could be trivial because the model is easy to accomplish this task only based on the knowledge context instead of the language context, which is more varied than knowledge context. To mitigate this, we discard neighbors of anchor nodes in 50\% of time during pre-training.

\subsection{Model Training}
CoLAKE is trained with cross-entropy loss. We use three classification heads to predict three types of nodes. In practice, however, the large number of entities brings challenges to training and predicting.

\paragraph{Mixed CPU-GPU Training.} Due to the large number of entities in KG, training the whole model in GPU is intractable. To handle this, we asynchronously update entity embeddings in CPU memory while keeping the rest of our model updated in GPU. In particular, we store and update entity embeddings in CPU memory which is shared among multiple trainer processes. During pre-training, the trainer processes read the entity embeddings from the shared CPU memory and write the gradients back to CPU. Our implementation is based on the distributed key-value store (KVStore) from Zheng et al.~\shortcite{Zheng2020DGLKE}.

\paragraph{Negative Sampling.} Applying the Softmax function to the huge number of entities is very time-consuming. CoLAKE uses negative sampling to conduct prediction for each entity over one positive entity and $k (k\ll n)$ negative entities instead of all $n$ entities in KG. Following Mikolov et al.~\shortcite{Mikolov2013word2vec}, we sample negative entities from the 3/4 powered entity frequency distribution.

\section{Experiments}
\label{sec:exp}
In this section, we present the details of pre-training and fine-tuning CoLAKE, and its experimental results on knowledge-driven tasks, knowledge probing tasks, and language understanding tasks.

\subsection{Pre-Training Data and Implementation Details}
CoLAKE uses English Wikipedia (2020/03/01)\footnote{\url{https://dumps.wikimedia.org/enwiki/}} as pre-training data and uses WikiExtractor\footnote{\url{https://github.com/attardi/wikiextractor}} to process the downloaded Wikipedia dump. We use Wikipedia anchors to align text to Wikidata5M~\cite{Wang2019KEPLER}, which is a newly proposed large-scale KG containing 21M fact triplets. We construct WK graphs as training samples and filter out graph samples without entity nodes and relation nodes. Finally, CoLAKE pre-trained the Transformer encoder along with 3,085,345 entity embeddings and 822 relation embeddings on 26M training samples.

The Transformer encoder of CoLAKE is initialized with RoBERTa$_\mathrm{BASE}$~\cite{Liu2019Roberta}. We use the implementation from HuggingFace's Transformer~\cite{Wolf2019HuggingFacesTS}. The entity embeddings and relation embeddings are initialized with the average of the RoBERTa$_\mathrm{BASE}$ BPE embeddings of entity and relation aliases provided by Wang et al.~\shortcite{Wang2019KEPLER}. AdamW with $\beta_1=0.9$, $\beta_2=0.98$ is used in pre-training. We train CoLAKE with the batch size of 2048 and the learning rate of 1e-4 for 1 epoch. For each anchor node, we sample $k=200$ negative entities. CoLAKE is trained on 8 32G NVIDIA V100 GPUs for 38 hours.

\subsection{Knowledge-Driven Tasks}
We first fine-tune and evaluate CoLAKE on knowledge-driven tasks. To annotate entities in the sentence, we use TAGME~\cite{Ferragina2010TAGME} to link mentions to entities in KGs. Instead of replacing the textual mention with its symbolic entity, we follow P{\"{o}}rner et al.~\shortcite{Porner2019LAMAUHN} and concatenate the two forms of tokens, e.g. Jean Mara \#\#is \texttt{Jean\_Marais}. Concretely, we conduct experiments on two knowledge-driven tasks: entity typing and relation extraction.

\paragraph{Entity Typing.} The entity typing task is to classify the semantic type of a given entity mention based on its surface form and context. We add two special tokens, \texttt{[ENT]} and \texttt{[/ENT]}, before and after the entity mentions to be classified and use the final representation of the \texttt{[CLS]} token as the feature to conduct classification\footnote{In the implementation of RoBERTa, the \texttt{[CLS]} token is replaced with \texttt{<s>}.}. We evaluate CoLAKE on Open Entity~\cite{Choi2018Ultra}. To compare with ERNIE, KnowBERT, and KEPLER, we adopt the same experiment setting which considers nine general types. To be consistent with previous work, we adopt micro precision, recall, and F1 score as evaluation metrics. The experimental results are shown in Table~\ref{tab:know}.

\paragraph{Relation Extraction.} The relation extraction task is to classify the relationship between two entities mentioned in a given sentence. During fine-tuning, we add four special tokens, \texttt{[HD]}, \texttt{[/HD]}, \texttt{[TL]} and \texttt{[/TL]} to identify the head entity and the tail entity. Also, we use the final representation of the \texttt{[CLS]} token as the feature to be fed into the classifier. We evaluate CoLAKE on FewRel~\cite{Han2018Fewrel} that is rearranged by Zhang et al.~\shortcite{Zhang2019ERNIE}. Since FewRel is built with Wikidata, we discard triplets in the FewRel test set from pre-training data to avoid information leakage. Following previous work, we report macro precision, recall and F1 score on FewRel. The experimental results can be found in Table~\ref{tab:know}.

\begin{table}[htbp]
  \centering\small
  \begin{tabular}{l|ccc|ccc}
    \toprule
    \multirow{2}{*}{Model}          & \multicolumn{3}{c|}{Open Entity}      & \multicolumn{3}{c}{FewRel}                 \\
                                    & P         & R         & F             & P            & R           & F             \\ \midrule
    BERT~\cite{Devlin2019BERT}      & 76.4      & 71.0      & 73.6          & 85.0         & 85.1        & 84.9          \\
    RoBERTa~\cite{Liu2019Roberta}   & 77.4      & 73.6      & 75.4          & 85.4         & 85.4        & 85.3          \\
    ERNIE~\cite{Zhang2019ERNIE}     & 78.4      & 72.9      & 75.6          & 88.5         & 88.4        & 88.3          \\
    KnowBERT~\cite{Peters2019Know}  & \textbf{78.6} & 73.7  & 76.1          & -            & -           & -             \\
    KEPLER~\cite{Wang2019KEPLER}    & 77.8      & 74.6      & 76.2          & -            & -           & -             \\
    E-BERT~\cite{Porner2019LAMAUHN} & -         & -         & -             & 88.6         & 88.5        & 88.5          \\ \midrule
    CoLAKE (Ours)                    & 77.0  & \textbf{75.7} & \textbf{76.4} & \textbf{90.6}& \textbf{90.6}& \textbf{90.5} \\ \bottomrule
    \end{tabular}
  \caption{Experimental results on Open Entity and FewRel.}
  \label{tab:know}
\end{table}

\subsection{Knowledge Probing}
LAMA (LAnguage Model Analysis) probe~\cite{Petroni2019LAMA} aims to measure factual knowledge stored in language models via cloze-style statement like: \emph{Dante was born in [MASK]}. Subsequently, a more "factual" subset of LAMA, LAMA-UHN~\cite{Porner2019LAMAUHN}, is constructed by filtering out easy-to-answer samples. We evaluate CoLAKE on these two probes and report the mean precision at one (P@1) macro-averaged over relations.

For fair comparision, we use the intersection of the vocabularies for all considered models and construct a common vocabulary of $\sim$18K case-sensitive tokens. In this experiment, considered models include ELMo~\cite{Peters2018Deep}, ELMo5.5B~\cite{Peters2018Deep}, BERT$_\mathrm{BASE}$~\cite{Devlin2019BERT}, RoBERTa$_\mathrm{BASE}$~\cite{Liu2019Roberta}, and K-Adapter~\cite{Wang2020Kadapter}.

\begin{table}[http]
  \centering\small
  \begin{tabular}{lcccccc}
    \toprule
    \multirow{2}{*}{Corpus} & \multicolumn{6}{c}{Pre-trained Models}                  \\
                            & ELMo & ELMo5.5B & BERT & RoBERTa & CoLAKE & K-Adapter$^*$     \\ \midrule
    LAMA-Google-RE          & 2.2  & 3.1     & 11.4 & 5.3      & 9.5   & 7.0      \\
    LAMA-UHN-Google-RE      & 2.3  & 2.7     & 5.7  & 2.2      & 4.9   & 3.7     \\ \midrule
    LAMA-T-REx              & 0.2  & 0.3     & 32.5 & 24.7     & 28.8  & 29.1     \\
    LAMA-UHN-T-REx          & 0.2  & 0.2     & 23.3 & 17.0     & 20.4  & 23.0     \\ \bottomrule
    \end{tabular}
  \caption{P@1 on LAMA and LAMA-UHN. $^*$K-Adapter is based on RoBERTa$_\mathrm{LARGE}$ while other Transformer-based LMs are of $\mathrm{BASE}$ size. Besides, K-Adapter uses a subset of T-REx as its training data, which may contribute to its superiority over CoLAKE on LAMA-T-REx and LAMA-UHN-T-REx.}
  \label{tab:lama}
\end{table}

The results of LAMA and LAMA-UHN are shown in Table~\ref{tab:lama}. It is worth noticing that BERT outperforms RoBERTa by a large margin. Wang et al.~\shortcite{Wang2020Kadapter} reported the same phenomenon in their paper. We conjecture that the main reason behind this is the larger and byte-level BPE vocabulary used by RoBERTa. Though, CoLAKE outperforms its baseline, RoBERTa$_\mathrm{BASE}$, by a significant margin. Besides, CoLAKE even improves over the $\mathrm{LARGE}$ size of model, K-Adapter, by 2.5\% and 1.2\% on LAMA-Google-RE and LAMA-UHN-Google-RE respectively.

\subsection{Language Understanding Tasks}
We also evaluate CoLAKE on the General Language Understanding Evaluation (GLUE)~\cite{Wang2019GLUE}, which provides a collection of diverse NLU tasks. Since these tasks require little factual knowledge, we attempt to explore whether CoLAKE degenerates the performance on these NLU tasks.

The experimental results on GLUE dev set are shown in Table~\ref{tab:glue}. CoLAKE is slightly degraded from RoBERTa but improves over KEPLER by 1.4\% on average. We conclude that CoLAKE is able to simultaneously model text and knowledge via the heterogeneous WK graph. In summary, our experiments demonstrate that CoLAKE significantly improves the performance on knowledge-required tasks and at the same time achieves comparable results on language understanding tasks.

\begin{table}[thbp]
  \centering\small
  \begin{tabular}{l|cccccccc|c}
  \toprule
  Model     & MNLI (m/mm) & QQP   & QNLI & SST-2 & CoLA & STS-B & MRPC & RTE  & AVG. \\ \midrule
  RoBERTa   & 87.5 / 87.3 & 91.9  & 92.8 & 94.8  & 63.6 & 91.2  & 90.2 & 78.7 & 86.4 \\
  KEPLER    & 87.2 / 86.5 & 91.5  & 92.4 & 94.4  & 62.3 & 89.4  & 89.3 & 70.8 & 84.9 \\
  CoLAKE     & 87.4 / 87.2 & 92.0  & 92.4 & 94.6  & 63.4 & 90.8  & 90.9 & 77.9 & 86.3 \\ \bottomrule
  \end{tabular}
  \caption{GLUE results on dev set. Both of KEPLER and CoLAKE are initialized with RoBERTa$_\mathrm{BASE}$.}
  \label{tab:glue}
  \end{table}

  \subsection{Word-Knowledge Graph Completion}
  Note that CoLAKE is essentially a GNN pre-trained on large-scale WK graphs, which makes it structure-aware and easy to generalize to unseen entities.

\begin{figure}[t]
    \centering
    \includegraphics[width=0.8\columnwidth]{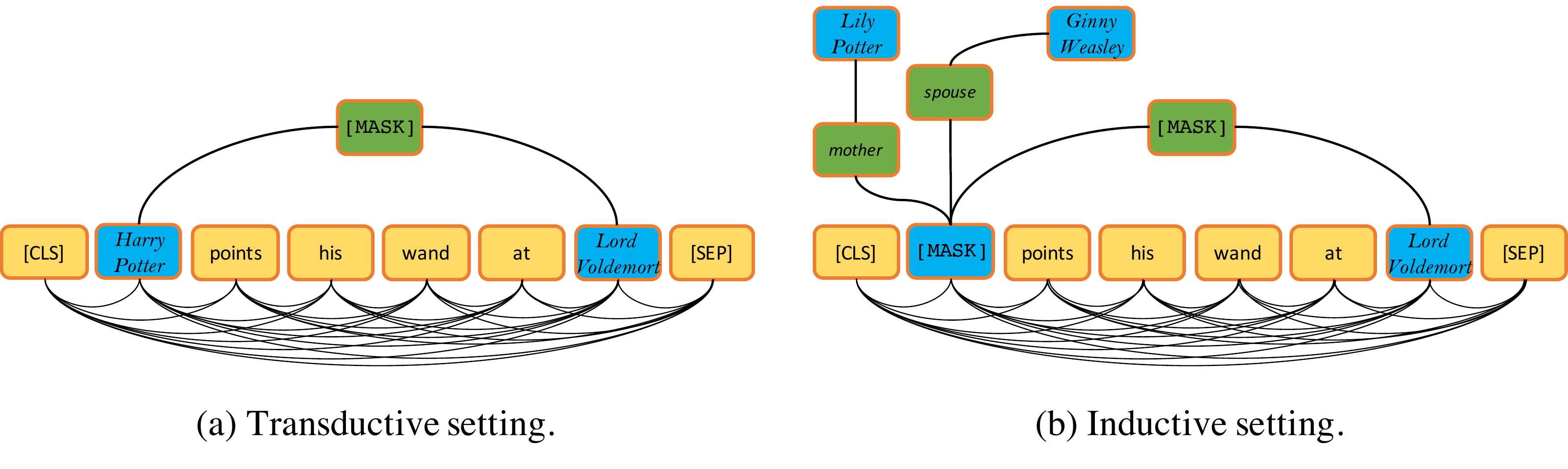}
    \caption{Illustration of the input word-knowledge graph for transductive setting and inductive setting. In transductive setting, \texttt{Harry\_Potter} and \texttt{Lord\_Voldemort} are seen during training. In inductive setting, \texttt{Harry\_Potter} is unknown but its neighboring entities are seen in training data.}
    \label{fig:anal}
  \end{figure}

  To probe CoLAKE's capability of modeling both structural and semantic features, we design a task named \emph{word-knowledge graph completion}. In particular, we use the FewRel test set to construct two experimental settings: transductive setting and inductive setting. In both settings, each sample provides a triplet $(h, r, t)$ and a sentence that expresses the triplet. The considered models are required to predict the relation $r$. For each sample in \textbf{transductive setting}, the two entities, $h$ and $t$, and their relation $r$ are seen in training phase. But the triplet $(h, r, t)$ has not appeared in the training data. We collect 10K samples from the FewRel test set to construct the transductive setting. For each sample in \textbf{inductive setting}, at least one entity is unseen during training. This setting requires the model to be inductive so that it can generalize to unseen entities. We collect 1K samples from the FewRel test set to construct the inductive setting. We directly evaluate CoLAKE in the two settings without further training on the FewRel training set. The forms of word-knowledge graph are depicted in Figure~\ref{fig:anal}. For inductive setting, we encourage CoLAKE to infer the unseen entity by aggregating messages from its neighbors.

  We take several well-known models for link prediction as our baselines\footnote{We did not compare with KEPLER~\cite{Wang2019KEPLER} since the authors did not release their data split and model yet.}. For transductive setting, we compare CoLAKE with four widely-used models, i.e. TransE~\cite{Bordes2013TransE}, DistMult~\cite{Yang2015DistMult}, ComplEx~\cite{Trouillon2016Complex}, and RotatE~\cite{Sun2019Rotate}. We use DGL-KE~\cite{Zheng2020DGLKE} to train the four baseline models on Wikidata5M\footnote{The triplets in FewRel test set are removed from Wikidata5M to avoid information leakage.}. For inductive setting, we take DKRL~\cite{Xie2016DKRL} as our baseline. As shown in Table~\ref{tab:anal}, CoLAKE outperforms other models by a large margin thanks to its capability of simultaneously utilizing structural knowledge and rich text semantics while traditional KE models can only handle structural knowledge. Besides, the inductive ability of CoLAKE is more realistic. Unlike DKRL and KEPLER, which generate entity embeddings from descriptions, CoLAKE generates entity embeddings based on their neighbors.

  \begin{table}[htbp]
    \centering\small
    \begin{tabular}{lccccc}
    \toprule
    Model                               & MR $\downarrow$ & MRR   & HITS@1 & HITS@3 & HITS@10 \\ \midrule
    \multicolumn{6}{c}{Transductive setting}                                        \\ \midrule
    TransE~\cite{Bordes2013TransE}      & 15.97 & 67.30 & 60.28  & 70.96  & 79.75   \\
    DistMult~\cite{Yang2015DistMult}    & 27.09 & 60.56 & 48.66  & 69.69  & 79.61   \\
    ComplEx~\cite{Trouillon2016Complex} & 26.73 & 61.09 & 49.80  & 70.64  & 79.78   \\
    RotatE~\cite{Sun2019Rotate}         & 30.36 & 70.90 & 64.74  & 74.89  & 81.05   \\
    CoLAKE                              & 2.03  & 82.48 & 72.14  & 92.19  & 98.58   \\ \midrule
    \multicolumn{6}{c}{Inductive setting}                                           \\ \midrule
    DKRL~\cite{Xie2016DKRL}             & 168.21& 8.18  & 5.03   & 7.28   & 14.13   \\
    CoLAKE                              & 31.01 & 28.10 & 15.69  & 30.28  & 58.05   \\ \bottomrule
    \end{tabular}
    \caption{The experimental results on word-knowledge graph completion task.}
    \label{tab:anal}
  \end{table}




\section{Conclusion}

In this paper, we propose CoLAKE to jointly learn contextualized representation for language and knowledge. We integrate the language context and knowledge context in a unified data structure, word-knowledge graph. The experimental results show the effectiveness of CoLAKE on knowledge-required tasks. Besides, to explore the potential application of the WK graph, we design a task named WK graph completion, which shows that CoLAKE is essentially a powerful GNN that is structure-aware and inductive. The surprisingly high performance on WK graph completion inspires the potential applications of WK graph, for example, (a) CoLAKE may help to denoise distantly annotated samples of relation extraction~\cite{Craven1999Constructing,Mintz2009Distant}, (b) CoLAKE can be used to measure the quality of graph-to-text templates~\cite{Davison2019Commonsense,Bouraoui2020Inducing} due to its capability of preserving the original graph structure. We leave these applications as future work.

\section*{Acknowledgements}

This work was supported by the National Natural Science Foundation of China (No. 61751201, 62022027 and 61976056), Shanghai Municipal Science and Technology Major Project (No. 2018SHZDZX01) and ZJLab.


\bibliographystyle{coling}
\bibliography{coling2020}

\end{document}